\documentclass[journal,10pt]{IEEEtran}
\usepackage{xcolor}
\usepackage{bbm}
\usepackage{amsmath}
\usepackage{acronym}
\usepackage{epsf}
\usepackage{times}
\usepackage{epsfig}
\usepackage{notoccite}
\usepackage{graphicx}
\usepackage{epstopdf}
\usepackage{pstricks}
\usepackage{amssymb}
\usepackage{amsxtra}
\usepackage{here}
\usepackage{rawfonts}
\usepackage{float}
\usepackage{times}
\usepackage{url}
\usepackage{cite}
\usepackage{caption}
\usepackage{subcaption}
\usepackage{algorithm}
\usepackage{algpseudocode}
\usepackage{blindtext}
\usepackage{enumitem}
\usepackage{xcolor,cite,etoolbox}
\usepackage{relsize}
\usepackage{lipsum}
\usepackage{graphicx}
\usepackage{tabularx}
\usepackage{xparse}
\usepackage{array}
\newcolumntype{P}[1]{>{\centering\arraybackslash}p{#1}}
\usepackage{mleftright}

\usepackage{mathtools}

\usepackage{graphics}
\usepackage{physics}
\usepackage{amssymb}
\usepackage{siunitx}
\usepackage{multicol}

\usepackage[nomain,acronym,shortcuts]{glossaries}
\makeglossaries

\usepackage{datetime}
\usepackage{amssymb}
\usepackage{subcaption}
\usepackage{verbatim}

\begin{document}

\title{Internet of Federated Digital Twins: Connecting Twins Beyond Borders for Society 5.0
\thanks{This research is funded by the NICT-JUNO \#22404, and by the U.S. National Science Foundation under Grant CNS-2210254.}
\thanks{T. Yu, Z. Li, and K. Sakaguchi are with the Department of Electrical and Electronic Engineering, Tokyo Institute of Technology, Tokyo, Japan (Emails: yutao@mobile.ee.titech.ac.jp, lizd@mobile.ee.titech.ac.jp, sakaguchi@mobile.ee.titech.ac.jp).}
\thanks{O. Hashash and W. Saad are with Wireless@VT, Bradley Department of Electrical and Computer Engineering, Virginia Tech, Arlington, VA, 22203, USA
(Emails: omarnh@vt.edu, walids@vt.edu).}
\thanks{M. Debbah is with KU 6G Research Center, Khalifa University of Science and Technology, P O Box 127788, Abu Dhabi, UAE (Email: merouane.debbah@ku.ac.ae) and also with CentraleSupelec, University Paris-Saclay, 91192 Gif-sur-Yvette, France.}}%

\author{Tao~Yu,~\IEEEmembership{Member,~IEEE,}
Zongdian~Li,~\IEEEmembership{Member,~IEEE,}
Omar~Hashash,~\IEEEmembership{Graduate~Student~Member,~IEEE,}
Kei~Sakaguchi,~\IEEEmembership{Senior~Member,~IEEE,}
Walid~Saad,~\IEEEmembership{Fellow,~IEEE,}
and~M{\'e}rouane~Debbah,~\IEEEmembership{Fellow,~IEEE}}%

\maketitle

%TC:ignore
\begin{abstract}
The concept of digital twin (DT), which enables the creation of a programmable, digital representation of physical systems, is expected to revolutionize future industries and will lie at the heart of the vision of a future smart society, namely, Society 5.0, in which high integration between cyber (digital) and physical spaces is exploited to bring economic and societal advancements. However, the success of such a DT-driven Society 5.0 requires a synergistic convergence of artificial intelligence and networking technologies into an integrated, programmable system that can coordinate DT networks to effectively deliver diverse Society 5.0 services. Prior works remain restricted to either qualitative study, simple analysis or software implementations of a single DT, and thus, they cannot provide the highly synergistic integration of digital and physical spaces as required by Society 5.0. In contrast, this paper envisions a novel concept of an \emph{Internet of Federated Digital Twins (IoFDT)} that holistically integrates heterogeneous and physically separated DTs representing different Society 5.0 services within a single framework and system. For this concept of IoFDT, we first introduce a hierarchical architecture that integrates federated DTs through horizontal and vertical interactions, bridging cyber and physical spaces to unlock new possibilities. Then, we discuss challenges of realizing IoFDT, highlighting the intricacies across communication, computing, and AI-native networks while also underscoring potential innovative solutions. Subsequently, we elaborate on the importance of the implementation of a unified IoFDT platform that integrates all technical components and orchestrates their interactions, emphasizing the necessity of practical experimental platforms with a focus on real-world applications in areas like smart mobility.
\end{abstract}
%TC:endignore

\section{Introduction}
Digital twins (DTs) are a transformative technology crafting faithful digital representations of physical systems, processes, and dynamics in cyber spaces. This application space enables DTs to span elements ranging within the Internet of Things (IoT) to intelligent transportation systems (ITS), healthcare, and intricate industrial systems across the whole product life-cycle, including design, manufacturing, distribution, and recycling \cite{1,2}. In essence, a fully realized DT is not just a static blueprint or simulation of a physical system, but a dynamic, high-precision, granular replica of a physical system, including its interactions with other physical system. Thus, DTs symbolize a harmonious blend of physical and cyber spaces, which departs from the traditional IoT concept, where interconnectivity is confined among physical objects and data flows mainly in one way from physical to cyber spaces. DTs will be an integral part of future applications ranging from manufacturing to agriculture and enabling the metaverse \cite{3}.

DTs will shape the future as a cornerstone of super smart society, dubbed Society 5.0 \cite{4}, where the intertwining of cyber and physical spaces drives economic and societal advancement across industries ranging from intelligent transportation to factory automation and robotics. The success of this DT-driven Society 5.0 depends on the seamlessly integration of physical and cyber spaces, ensure precise coordination among DTs and enable them to work together effectively to deliver diverse Society 5.0 services.
DTs, working in synergy, streamline end-to-end (E2E) operations demanding synchronization across multiple Society 5.0 services, such as smart mobility and automated manufacturing. Hence, DTs can establish interconnections and form cooperative clusters in cyber space, collectively modeling and analyzing their corresponding physical systems at varying granularities. Therefore, as a milestone towards enabling Society 5.0, it is necessary to understand challenges of synergistically interconnecting multiple DTs over wireless networks rather than considering each DT in isolation.

Recent research exploring synergies between DTs and wireless systems falls into three main categories. The first category (e.g., in \cite{5}), constituting a significant portion of existing research, leverages DTs to create digital representations for wireless systems (e.g., 5G) to enhance network management. While these works are beneficial for wireless systems, they neglect the wireless systems's role in connecting and synchronizing multiple DTs. The second category (e.g., in \cite{6}) studies the deployment of a single, isolated DT within wireless systems, focusing on individual networking and learning schemes. Clearly, these works fall short in extrapolating these network mechanisms for a complex system of interconnected DTs. Finally, some works (e.g., in \cite{7}) focus on virtual representation of twins, neglecting network constraints and primarily tackling software development hurdles. Therefore, these approaches fall short of realizing comprehensive and coordinated E2E DT services in Society 5.0, as current DT technologies lack required features, such as scalability and interconnectivity, that are indispensable in systems of interconnected DTs. Comprehensive reviews on DTs can be found in \cite{1,2}.
While existing works have explored broader aspects of DTs, they lack a focused exploration on complex integration and coordination of multiple DTs across various physical systems with seamless interoperability, a gap that this paper aims to address.

Building on this identified gap, this paper introduces and defines the \emph{Internet of Federated Digital Twins (IoFDT)}, a novel concept designed to enhance interconnections, coordination, and cooperation among DTs across diverse systems while addressing critical performance metrics such as efficiency, scalability, and latency. In essence, the IoFDT plays a crucial role and constitutes the backbone in realizing Society 5.0 by interconnecting a comprehensive, federated network of diverse DTs and Society 5.0 services. Each DT within an IoFDT in the vision of Society 5.0 is designed for a specific system or service, with federation occurring not only among homogeneous DTs, but in broader heterogeneous systems. Moreover, the IoFDT architecture employs dynamic resource allocation and modular services that can adjust system demands, ensuring efficient scalability and performance integrity of the growing DT network.

\begin{figure}[t]
\centering
\includegraphics[width=\linewidth]{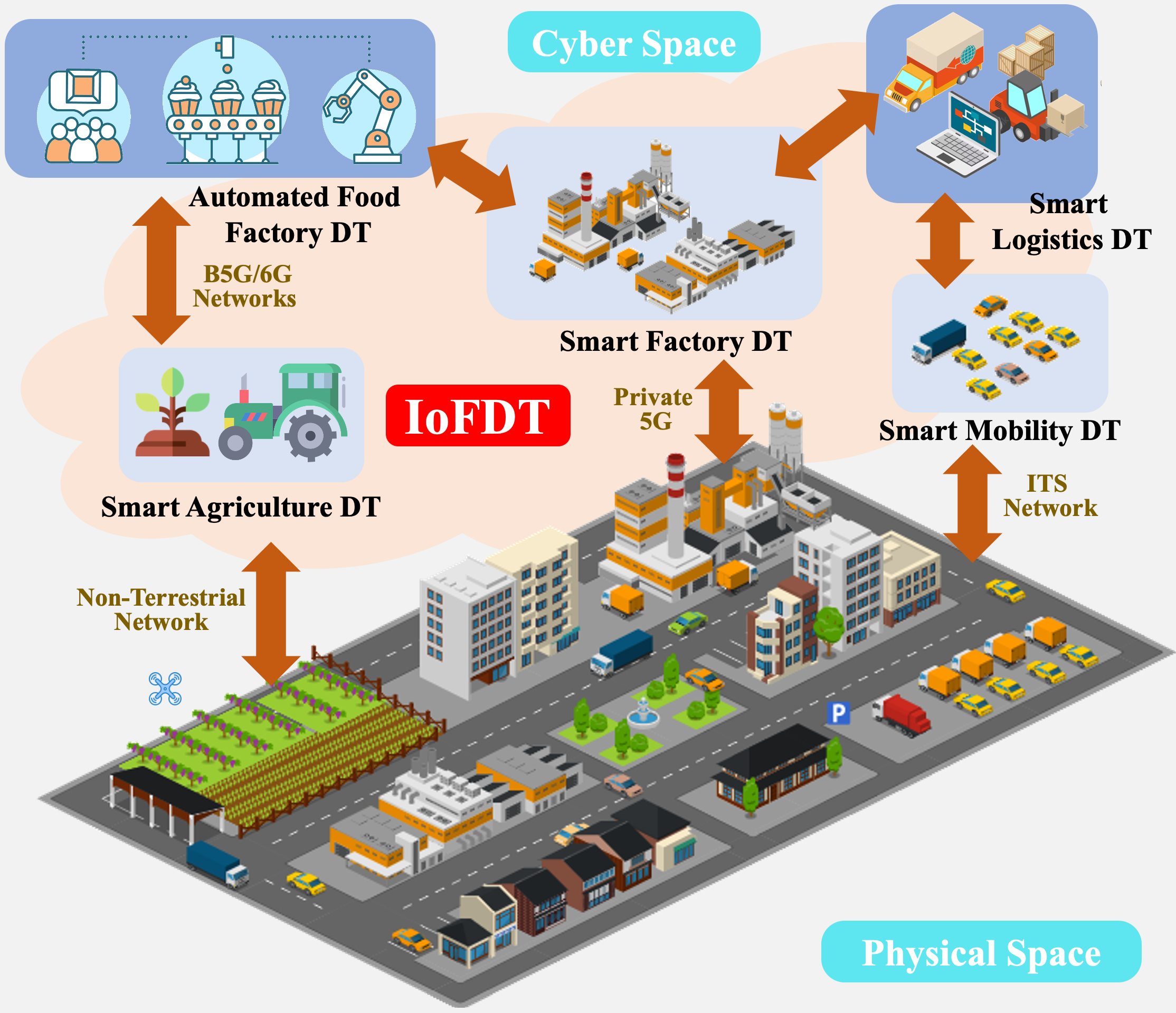}
\caption{\small An example of the vision of Internet of Federated Digital Twins over wireless networks.}
\label{fig:model}
\end{figure}

This creates a complex federation of interconnected DTs across various domains in Society 5.0, all linked over wireless systems. In this framework, a federated DT leverages sliced component functions, including sensors, networks, and computing resources integrated within an IoFDT platform. The IoFDT coordinates multiple federated DTs and integrates physically separate, heterogeneous DTs beyond their physical borders. Fig.~\ref{fig:model} illustrates an example of the IoFDT with several DTs in the vision of Society 5.0. In this example, the connection between the factory DT and agriculture DT in cyber space enables a food processing DT, which manages agricultural products from farms and food processing chain in factories in cyber space. Such integration optimizes harvest schedules, production plans, quality control, and E2E traceability.
Other examples requiring an interconnected IoFDT include: city energy management by connecting power plant DT with user behavior DT; smart supply management by linking logistics DT and manufacturing DT; precision construction by associating construction site DT with construction machinery DT; and smart preventive healthcare by connecting user lifestyle DT to medical system DT. The IoFDT significantly departs from conventional DT systems relying on isolated, independent DTs, and could provide a backbone for Society 5.0 to revolutionize future societies and deliver E2E smart services through the seamless integration of cyber and physical spaces.

Towards enabling the IoFDT, a confluence of AI, networking, computing, and implementation is essential, which introduces several novel challenges that should be addressed. This paper's key contributions are summarized as follows:

\begin{itemize}

\item We propose a novel IoFDT framework architecture marked by its intricate hierarchical structure of DTs that are not just interconnected but federated while bridging cyber and physical spaces. This architecture leverages multi-layered insights, enables operational, systemic, and predictive analytics from individual DTs to the collective federation, and thereby enhances decision-making, strategic foresight, and operation in the IoFDT ecosystem.

\item We identify multifaceted challenges at the crossroad of communications, artificial intelligence (AI), and computing. We put forth potential, novel solutions in realizing the IoFDT at enhancing synchronization, cross-layer networking, real-time computing, scalability, AI learning, AI-native networks, and generalizability of twins.

\item Transitioning from theory to practice, we outline the implementation of a unified experimental platform integrating DT, network, and computing orchestrators to facilitate seamless interactions among diverse DTs. We emphasize real-world testing to validate theoretical models and introduce a comprehensive proof-of-concept (PoC) implementation, initially focused on smart mobility to demonstrate the IoFDT's feasibility and merit.

\end{itemize}

The rest of this paper is organized as follows. Section II presents the overall IoFDT framework. Section III introduces enablers and challenges for IoFDT across networking, AI, and computing. Section IV introduces an experimental IoFDT platform with experimental applications. Section V concludes with future recommendations towards enabling IoFDT.

\section{IoFDT: From Individual DTs to Hierarchical, Federated DT Networks}
The IoFDT architecture features a network of interconnected, federated DTs, as illustrated in Fig.~\ref{fig:iofdt_dts}. This topology overcomes limitations inherent in standalone DT systems, such as lack of interconnectivity, inability to extract multi-layered insights, limited adaptability, and the absence of a collaborative ecosystem.

At the center of IoFDT lies a series of DTs, which serve as precise digital representations of physical objects or processes, bridging cyber and physical spaces. These DTs are strategically arranged in a tiered structure, with each level reflecting intricacies of DTs' functionalities and degrees of interaction with their physical counterparts. This hierarchy enables the extraction of rich, multi-layered insights, such as operational details from an irrigation machine DT, systemic understanding of the irrigation system's impact on crop growth, and predictive insights on planting cycles and pest control strategies from collective data of farm DTs. DTs, diverse as systems and services they represent, are not individual but intrinsically linked through a federation. This DT federation is more than a mere technical connection. Instead, it acts as a unifying, logical thread that weaves together distinct DTs, creating a rich tapestry of interactions and information/knowledge exchange. This shared network establishes a collaborative ecosystem within the IoFDT. Participants in this network, representing various DTs, actively exchange real-time data, performance metrics, and AI-derived insights. This continuous exchange not only leverages existing knowledge, but integrates fresh data, refining system's collective intelligence and adaptability. The DT federation fosters seamless data synchronization and knowledge sharing among DTs. For instance, a DT representing an irrigation system might adjust water distribution based on insights from another DT monitoring soil moisture. As DTs interact, they fine-tune their algorithms, ensuring their digital representations remain aligned with real-world conditions they twin. The IoFDT framework categorizes interactions into two types: horizontal interactions among DTs of similar hierarchies or functionalities; and vertical interactions among DTs across different hierarchical levels, bridging granular processes to overarching system dynamics. Next, we describe the horizontal and vertical splits in the IoFDT:

\begin{figure}
\centering
\includegraphics[width=\linewidth]{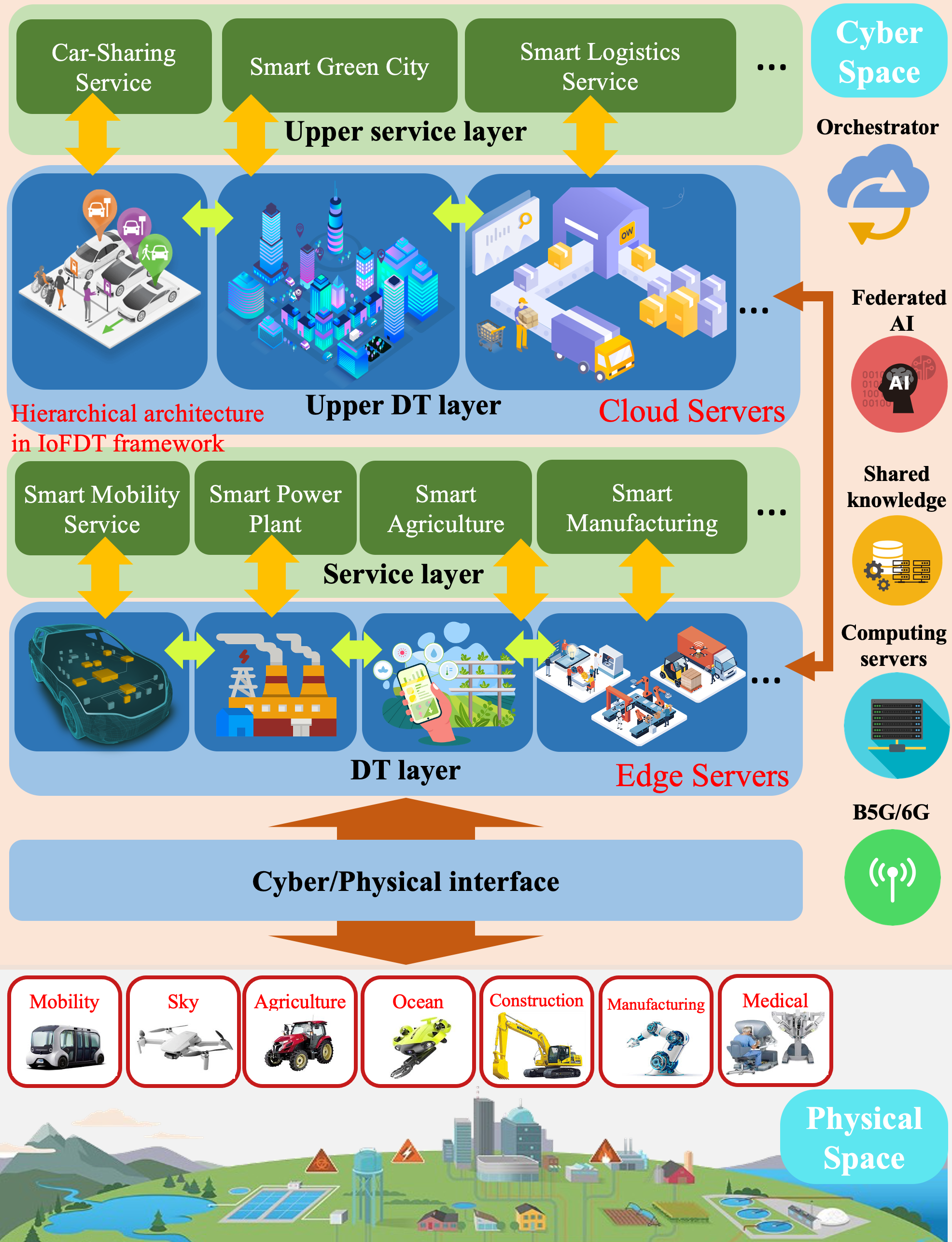}
\caption{\small Federated DTs and associated platform for smart services in IoFDT.}
\label{fig:iofdt_dts}
\end{figure}

\subsection{Horizontal Split} 
Horizontal interactions within the same DT levels and clusters in the hierarchy are fundamental to the IoFDT for enhancing collaboration, learning, and growth among geographically dispersed DTs. For example, consider the following scenario: A DT provides real-time crop yields and readiness information to a logistics DT, which plans optimal routes for fresh product transportation. Simultaneously, a supermarket chain DT integrates this information to manage inventory and plan their supply based on fresh product arrival. Meanwhile, planting or harvest plans on farms will be adjusted according to supermarket demands. This interaction is not only about data exchange, but constitutes continuous mutual learning and growth. Horizontal interaction is strengthened by sophisticated frameworks of real-time data sharing, advanced analytics, and AI-driven insights. It enables DTs to catch and understand other DTs through the IoFDT. Furthermore, when DTs, either homogeneous or heterogeneous, form functional clusters, the horizontal interactions also facilitate communication across DT clusters. In this interconnected, federated ecosystem, geographical distances and different segments become inconsequential. The IoFDT effectively shrinks the physical world, by bringing diverse twins closer together in a shared cyber space.

\subsection{Vertical Split}
The IoFDT also includes vertical interactions that bind DTs at higher hierarchical levels with those operating at more granular levels. While lower-layer DTs offer granular, real-time data and insights into specific physical systems, upper-layer DTs aggregate and contextualize this information, providing a holistic view for strategic decision-making. For instance, in a smart city, individual DTs might monitor specific services like traffic, energy, or waste management, with each providing specific insights into its respective physical system. An overarching city management DT, positioned at a higher hierarchical level, integrates these insights for decisions on overall resource allocation, emergency response, and long-term urban planning. Similarly, in a hospital, while individual DTs focus on specific departments or patient data, a high-tier healthcare system DT combines these insights to enhance patient care coordination to optimize resource distribution and refine overall hospital operations. This vertical interaction ensures not only that ground-level detailed insights are considered, but also that higher-level decisions are effectively disseminated, to maintain system-wide coherence and efficiency.

Having delineated the IoFDT's foundational framework, it is essential to address the underlying technical challenges and enablers over wireless systems. The intricacies and potential solutions at the intersection of communication, AI, and computing will be detailed next.

\section{Challenges and Enablers for IoFDT: Wireless, Learning, and Computing}
Despite IoFDT's benefits, realizing this vision still faces many challenges.

\subsection{Communications and Networking Challenges}

\subsubsection{Synchronization over the network}
Creating and federating multiple DTs requires real-time data accumulation from a multitude of physical objects and processes across vast distances. Meanwhile, modifications in cyber space must be reflected in almost real-time on the physical counterpart, ensuring DTs must be \emph{synchronized} with the physical space \cite{8}. The twin mandates of IoFDT, i.e., real-time data collection and actuation, present challenges, because they require real-time data transmission between devices of various physical systems and edge/cloud servers through complex, heterogeneous networks. It is also crucial to synchronize interactions between multiple DTs, each representing different aspects or components of a larger system. This further requires maintaining synchronization of contextual data relevance and operational coherence across DTs. Such demands could strain existing communication systems, such as 5G/6G, while requiring them to constantly adjust to dynamics and requirements of physical systems, ensuring seamless coordination between various DTs and physical processes. To address synchronization challenges, dynamic network slicing can be developed whereby network resources are adaptively allocated based on each DT's requirements. This approach, illustrated in Fig.~\ref{fig:slicing}, optimizes resource allocation within each slice according to the context and industry-dependent attributes of the physical system. Evolutionary reinforcement learning (ERL)\cite{9}, combining reinforcement learning (RL) with evolutionary algorithms, could be employed to evolve slicing policies over time to meet DT's dynamic requirements. Also, here one can exploit semantic communications to push forward less data and facilitate DT synchronization~\cite{10}.

\subsubsection{Cross-layer networking between DTs}

\begin{figure}
\centering
\includegraphics[width=\linewidth]{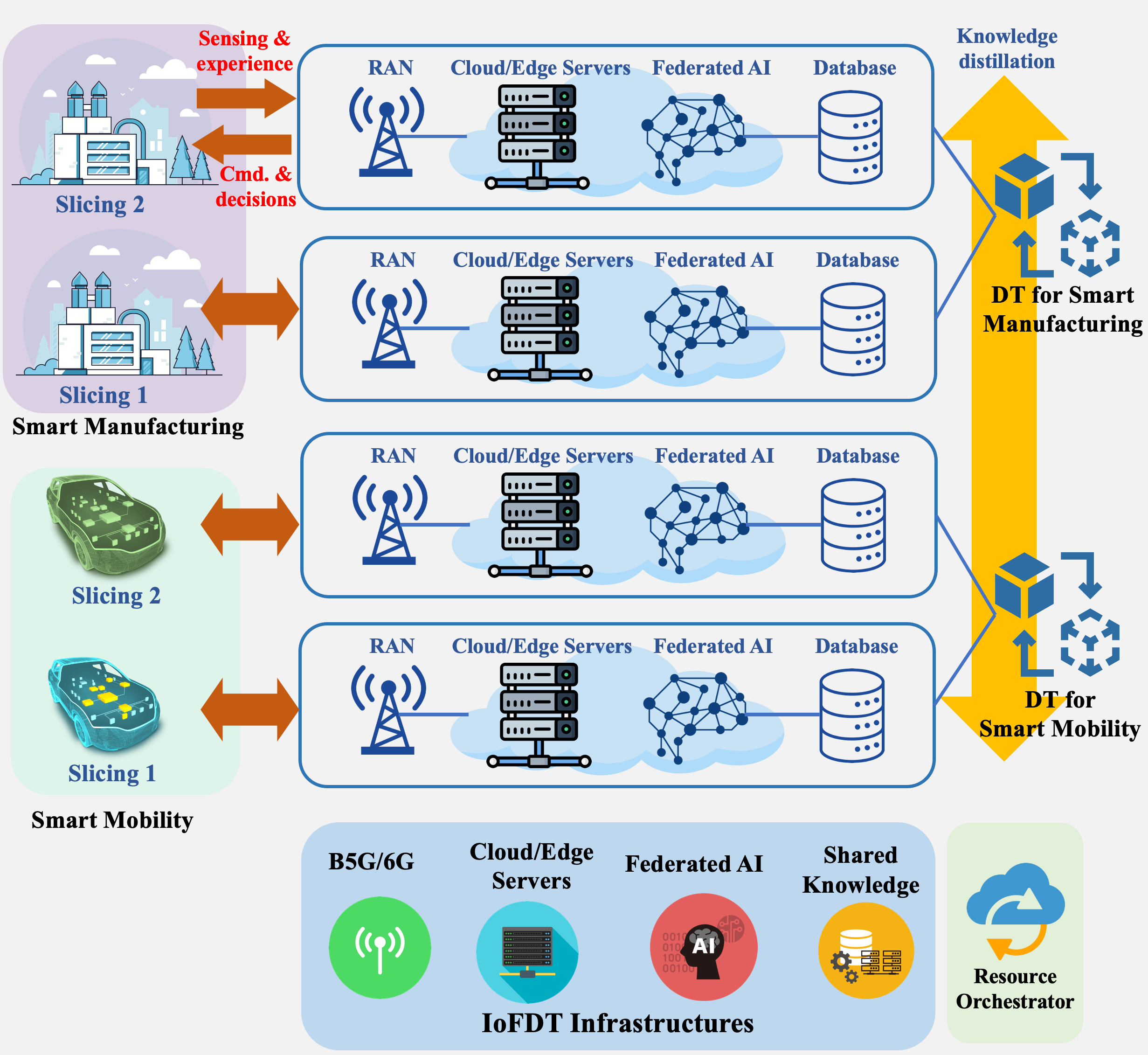}
\caption{\small Resources slicing and orchestration for DTs in IoFDT}
\label{fig:slicing}
\end{figure}

To harness DTs' benefits and realize IoFDT for Society 5.0, it is necessary to address a series of unique, cross-layer networking challenges inherent in IoFDT. Unlike conventional networks, IoFDT requires seamless interaction between various networking layers to support complex functionalities of federated DTs. Main challenges include managing asynchronous communications across different layers and DTs, ensuring data integrity across multiple network protocols, and orchestrating responsive information flows that can adapt to the rapidly changing states of physical entities represented by DTs. These challenges are magnified by diverse natures of devices involved, each with different communication protocols and data formats, necessitating a unified coherent network structure. Real-time analytics and decision-making processes based on multi-layer contextual information further increase complexity, requiring more intelligent and flexible network design and management. The tailored solution for these challenges involves network slicing, as shown in Fig.~\ref{fig:slicing}, to discern and dynamically accommodate distinct communication and processing requirements at each network layer. This includes the provisioning of network slices with specific cross-layer functionalities like protocol translation, data prioritization, and layer-specific security measures. Moreover, by utilizing a combination of collaborative reinforcement learning (CRL) and transfer learning, the network can continuously learn from data flow patterns and adjust slices in real-time, ensuring optimal cross-layer communication responsive to each DT's contextual needs. Such an approach enables dynamic and intelligent cross-layer networking that effectively sustains the complex operations of IoFDT to ensure that all network layers work collaboratively for the DT federation.

\subsection{IoFDT Computing Challenges}

\subsubsection{Real-time computing for data collection and actuation}
Real-time computing in the IoFDT is essential for effectively coordinating multiple DTs with one another and with their dynamic physical systems. This requires not only efficient architectural design, but agile resource allocation strategies for rapid processing of complex datasets on distributed edge servers, which often have limitations in computing power. The challenges in real-time computing are multifaceted. Latency issues arise from data transmission and processing across distributed computational nodes, and any delay can degrade decision-making. Edge servers must process large-scale data inflows, potentially causing bottlenecks. The computational infrastructure and resources require high flexibility, scaling to meet DTs' real-time needs. Consider smart mobility services in which autonomous vehicles' safety and efficiency hinge on immediate sensor data processing and communication with surrounding infrastructures. Any lag in these processes could result in outdated decisions, risking safety and operational inefficiencies. Moreover, when multiple DTs interact, the challenge escalates as each DT requires data from others in real-time to make informed decisions. This interdependence necessitates not only rapid data sharing but synchronized data processing and analysis across DTs, which demands an advanced level of computational orchestration for system-wide coherence and efficiency. To overcome these challenges, a potential solution is implementing edge computing architectures with specialized accelerators for data-intensive tasks, such as FPGAs and GPUs, which can significantly enhance real-time analytics processing at edges. Employing predictive resource allocation strategies utilizing machine learning models to forecast demand and preemptively adjust resource distribution can improve DTs' responsiveness. This predictive approach, combined with flexible edge computing resources, would minimize latency and enable more effective real-time computation and actuation for tasks that are critical to DTs' operation and synchronization, such as immediate traffic re-routing in connected vehicle networks.

\subsubsection{Scalability}
The scalability of the IoFDT infrastructure is a key challenge, because the IoFDT encompasses diverse, geographically dispersed DTs representing increasingly complex and dynamic physical systems. As DTs' number and complexity grow, computational demands and data volumes in IoFDT also grow, necessitating a distributed computing network capable of dynamic adjustment and scaling. Key scalability challenges include developing a computing architecture ensuring high-throughput data processing and rapid elasticity for growing and fluctuating workloads, while also being resilient to node failures to maintain operational integrity and system performance. A potential solution to these challenges lies in advanced distributed computing networks with dynamic adaptive load balancing, auto-scaling, and resource provisioning responding to DTs' increasing computational demands. Furthermore, spectral graph theory offers a powerful tool for modeling the complex network of DT interconnections. By treating the network as a graph, spectral analysis enables us to identify the interactivity degrees of DTs and computational nodes in IoFDT. Such insight allows for optimized resource allocation, and thereby can optimize network performance and reduce computational overhead. Essentially, spectral graph theory also aids in streamlining communication routes and prioritizing resource distribution to enhance overall efficiency and reduce latency in IoFDT.

\subsection{AI Challenges}

\subsubsection{AI Models for IoFDT Design}
The IoFDT requires new AI frameworks for interconnected and dynamic DTs evolving with data and real-time actuation. At its core, each DT is an AI model that continuously updates its physical space twin, and this is where AI is leveraged to create DTs and the entire IoFDT system. Learning within IoFDT must consider networking constraints while accurately twinning distributed physical processes across various dimensions. To fulfill network-aware AI twinning, AI must be adept at creating context-aware twin models that adapt at data inflows, and extract complex interrelations across diverse, heterogeneous systems. These frameworks must encapsulate the capability for multidimensional representations that accurately reflect physical processes and inform network design within system-wide communication and computing limitations. Moreover, compared to standalone DTs, generating and leveraging synthetic data for IoFDT to simulate hypothetical yet plausible scenarios pose greater challenges, as clusters of federated DTs involve increased complexity of dynamic, diverse, interconnected systems. \emph{Continual graph neural network (CGNN)}\cite{11}, combining benefits of continual learning (CL) with graph neural network (GNN), is a candidate promising to address these demands by adapting to temporal dynamics and complex interactions inherent in IoFDT. On the one hand, CL provides an agile incremental learning method with swift model updates~\cite{12}. On the other hand, GNNs present an effective method to model the interrelations between the multi-level DT elements. For instance, in smart manufacturing, IoFDT can integrate various DTs such as machinery, warehouse, and logistics. CGNNs are utilized to continuously learn and update relationships based on their complex data interactions and dependencies, thereby optimizing entire production process. When GNNs are not directly applicable, knowledge graphs can be employed to articulate and structure the non-graph-structured relationships.

\subsubsection{AI-Native Networks for IoFDT}
In parallel, the IoFDT necessitates AI-native communication networks to manage the autonomous coordination, synchronization, and connectivity of various federated DTs. These networks form the IoFDT's backbone, and enable not only the information flow but the actuation back to physical systems. AI techniques here must ensure seamless operation within communication networks in the IoFDT, address data's distributed nature across multiple devices, and facilitate real-time responses. This requires sophisticated orchestrations of network resources that are aware of the context and constraints in the IoFDT, while maintaining harmonious synchronization between physical and digital twins for optimized functionality. To address these challenges, we envision integrating adaptive learning algorithms that not only keep DTs updated but also enable real-time prediction and actuation for continuous synchronization between physical and digital twins in a scalable manner. Additionally, grounding AI-native networks in causal techniques, e.g., causal reasoning, can deepen network behavior understanding, leading to improved explainability, generalizability, and sustainability of the network operations, which enhances network performance for dynamic and interconnected DTs management~\cite{13}.

\subsubsection{Generalizability of Twins}
Moreover, AI systems within IoFDT must be able to generalize across new, previously unseen learning tasks in real-time. Such adaptability is critical as federated DTs constantly face novel situations and challenges. IoFDT AI frameworks must swiftly and accurately adjust learning algorithms and models in response to emerging tasks to improve federated DTs' continuous development and refinement. Leveraging advanced meta-learning and domain adaption techniques could serve as a viable solution, and the integration of these techniques could significantly enhance DTs' generalizability and evolutionary pace within the IoFDT ecosystem.

\section{Practical Implementation and Realization of IoFDT: Platform and Experimental Field}
In addition to theoretical challenges outlined in Section III, a unified, experimental IoFDT platform is needed to blend all technical elements and orchestrate their interactions. Indeed, experiments with practical Society 5.0 services in IoFDT on a fully functional experimental platform are indispensable to validate fundamental research and generate testing data for system design refinement.

\subsection{Unified IoFDT Platform to Integrate All Elements}

We propose an IoFDT platform, as a universal integrator and orchestrator, to unify all elements in IoFDT and orchestrate seamless interactions among DTs. To realize such a platform, three functional modules become essential: the DT orchestrator, network orchestrator, and computing orchestrator.

\textit{DT Orchestrator}:
This orchestrator manages DTs throughout life-cycle, including storage, creation, distribution, operation, and interaction. Central to this module are repositories of reusable DT functions and knowledge bases (AI models). For new applications and associated DTs, instead of starting from the ground up, they leverage pre-existing function modules and knowledge, accelerating DT development and implementation. The DT orchestrator distributes and deploys new DTs to optimized edge or cloud servers based on application scenarios and QoS requirements. It orchestrates DT interactions in the IoFDT, including data sharing and learning through DT federalization, data format/protocols coordination, data desensitization, and knowledge base management. In this containerization approach, the orchestrator assembles components to create DTs. These components, along with function modules, dependencies, configurations, AI models, and desensitized data, are then packaged into containers using tools such as Docker and containerd. They are then deployed and run by tools such as Kubernetes and Docker Swarm. The DT orchestrator, with a global system view, also decides the federation participation of new DTs, considering function, service, and resources. A continuous integration/continuous deployment (CI/CD) pipeline using automation tools like Jenkins ensures this process for each DT creation or update. Middleware is also necessary for data interoperability between networks, DTs, and devices using different data formats/protocols.

\textit{Network Orchestrator}:
This orchestrator manages the data flow and resources in IoFDT by allocating network resources, managing diverse communication protocols, and maintaining network QoS for each DT to collect data, process it, and send the input back to the physical space in real-time. Software-defined networking (SDN) and network function virtualization (NFV) can be employed for dynamic and programmable networks through an IoFDT network control plane. To orchestrate IoFDT networks and enable access to customized virtual network functions (VNFs) for each DT, SDN controllers provide a global, real-time, and unified view of resources and programmable interfaces to manipulate data flows. In practical implementation, SDN frameworks, such as OpenDaylight, Open Network Operating System, and Ryu, can be leveraged to abstract resources in networks, supporting the dynamic slicing for diverse DTs. RAN and core network functions are transformed into VNFs and managed by open platforms, such as Open RAN (O-RAN). An interface between the chosen SDN controller and the VNF management platform is necessary to monitor the status of VNFs. By implementing this composition, the IoFDT network orchestrator maps the resources required by DT to the specific physical infrastructures, controlling data flows to ensure seamless connectivity.

\textit{Computing Orchestrator}:
The computing orchestrator manages computing resources, both cloud and edge, to ensure that each DT can process the data and learn the physical space in real-time. It involves such as resource monitoring, workload balancing, dynamic resource allocation, and edge-cloud coordination. Such orchestration must be in real-time while constantly adapting to DTs' dynamic needs and the fluctuating availability of resources. Practically, each DT application can be packaged into a unit, such as a Pod in Kubernetes comprising DT and knowledge base containers. It can be deployed in a cloud-edge server computing cluster. The orchestrator, typically at the cloud server, monitors real-time resource usage in the cluster and DT demands, including memory, CPU, GPU, and storage, as assessed by tools such as Kubelet. With global insights and perspective, the orchestrator optimally allocates computing resources, ensuring real-time processing capabilities for DTs.

\subsection{Experimental Platform for IoFDT}

\begin{figure*}
\centering
\includegraphics[width=0.9\linewidth]{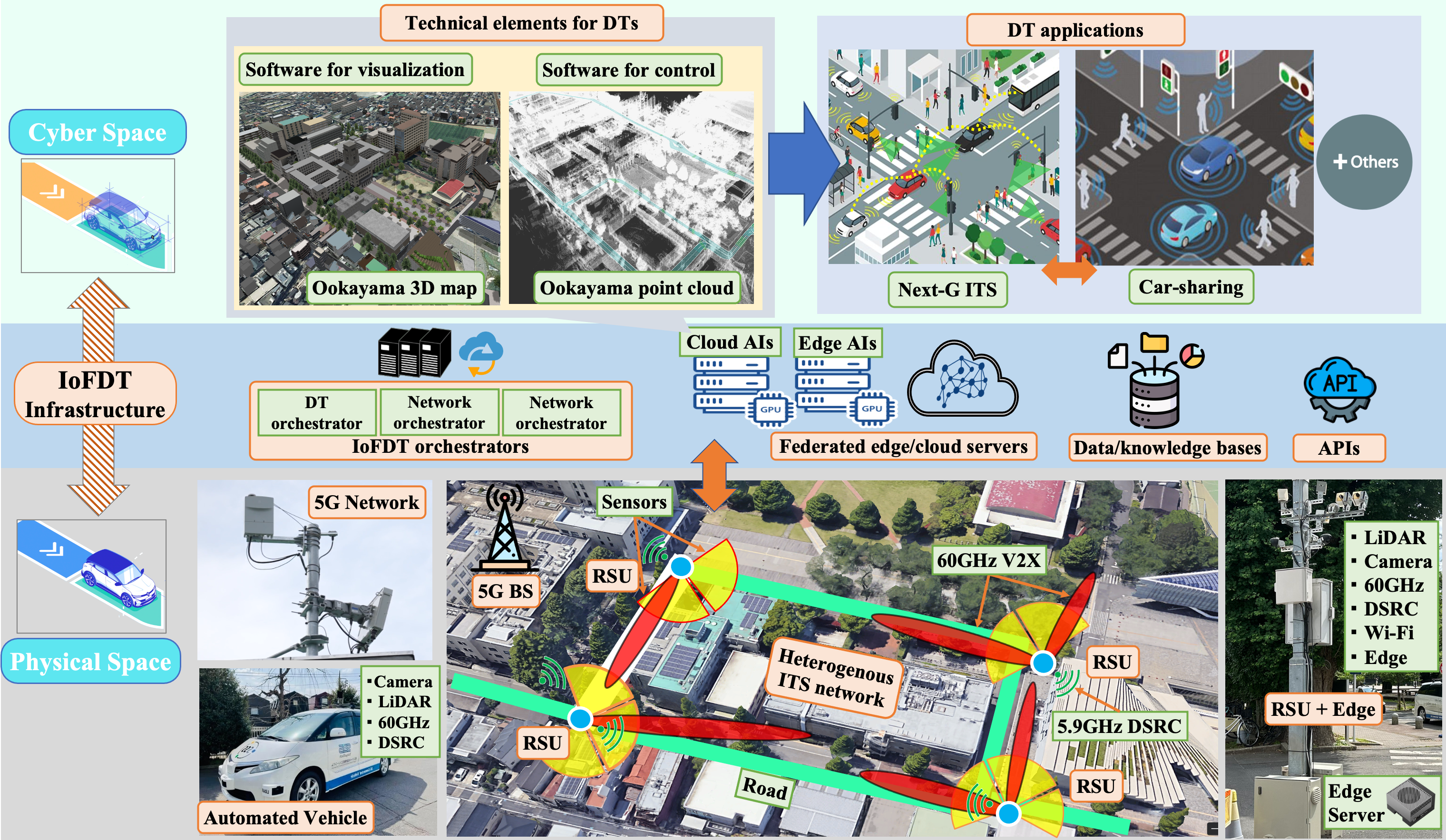}
\caption{\small Overview of IoFDT PoC and experimental field in Tokyo Tech}
\label{fig:poc}
\end{figure*}

An IoFDT experimental platform needs to be approached from two aspects.
\begin{itemize}

\item Establishing a comprehensive experimental PoC for IoFDT with interconnected federated DTs to support all its key features of learning data and running DTs;

\item Implementing a system-level PoC development and implementation, including hardware/infrastructure and software/algorithms for system demonstration and performance evaluation.

\end{itemize}

Fig.~\ref{fig:poc} shows a proposed architecture of an IoFDT PoC implementation with physical systems, network infrastructure, and DTs. It is an open, programmable, and universal platform, constructed with state-of-the-art, programmable sensing, communication, and computing hardware (e.g., LiDARs, 5G NR networks/ITS networks, MEC/cloud servers), and developer-friendly open-source software (e.g., Open Daylight, Robot Operating System, and TensorFlow/PyTorch). The experimental integrated IoFDT platform unifies cyber and physical spaces and satisfy the practical requirements of diverse applications. Given the platform's emphasis on mobility as an example, we developed a Smart Mobility Research \& Education Field in Tokyo Tech, as shown in Fig.~\ref{fig:poc}, with state-of-the-art infrastructures, e.g., 5G NR network, millimeter (mmWave) vehicular networks (backhaul/access)\cite{14}, roadside units (RSUs) and connected automated driving vehicles (CAVs) as a cornerstone of PoC implementation for IoFDT. 

\begin{figure}
\centering
\includegraphics[width=\linewidth]{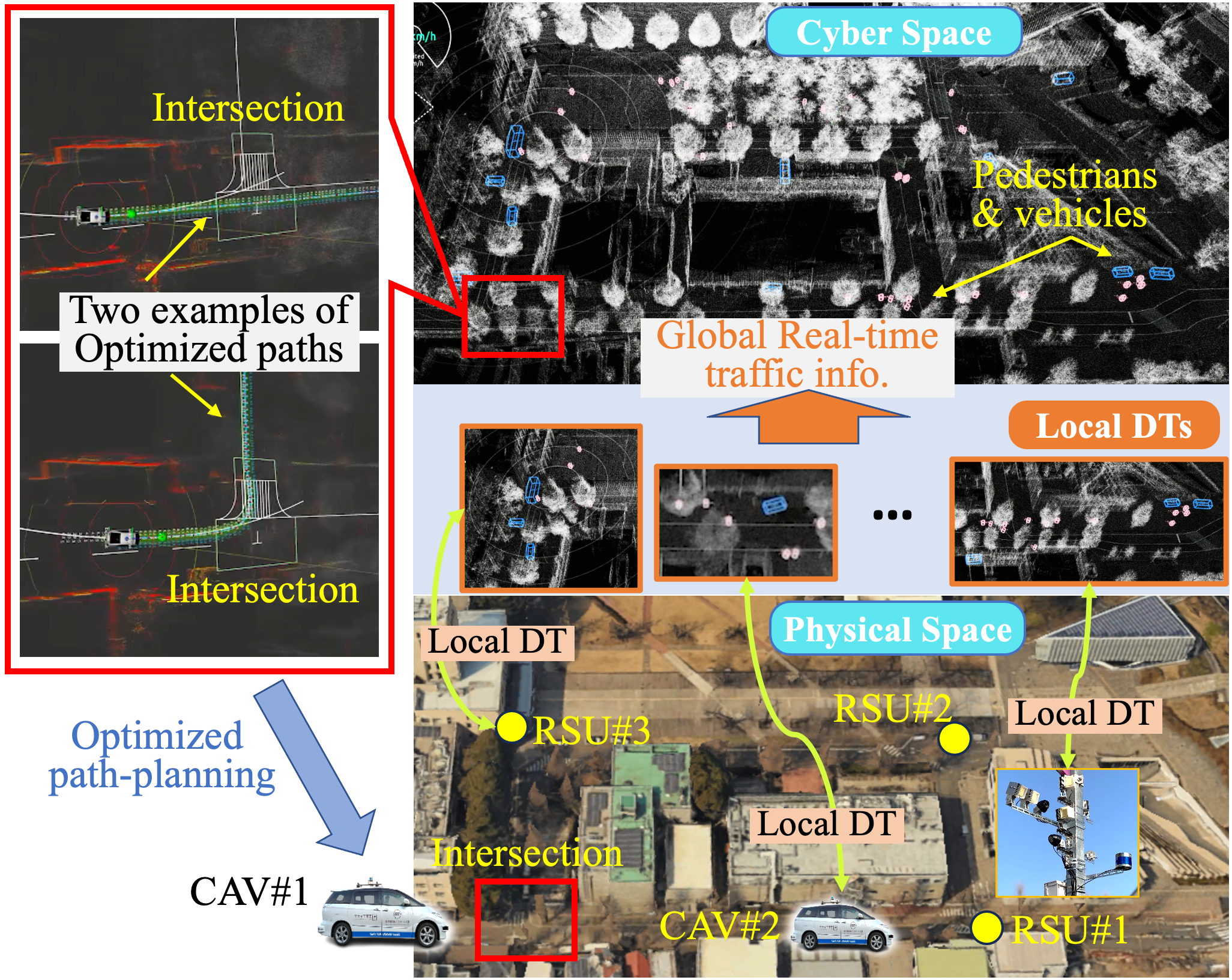}
\caption{\small Smart mobility DT and DT-based path optimization.}
\label{fig:demo}
\end{figure}

PoC trials demonstrated the system functions and enhanced vehicle perception, assisting safer automated driving. 
For example, as shown in Fig.~\ref{fig:demo}, we have developed a smart mobility DT that fuses traffic information detected, processed, and then mapped into cyberspace at RSUs and CAVs\cite{15}. This serves as an initial experimental implementation of integrating multiple local DTs to provide insights on the global traffic situation. A PoC application of path planning for CAVs was also conducted, in which optimized paths are calculated for CAVs based on real-time global traffic information in DT, and sent to CAV via vehicular network. The CAV then adjusts its path accordingly. A demo video of this PoC is available\footnote{https://youtu.be/Y3XXlNIRmXI}. The IoFDT PoC platform will be built on these preliminary experimental works as well as extensive works on AI, communication networks, computing networks, and platforms discussed in previous sections. We are also developing and implementing diverse smart mobility prototype applications on the IoFDT PoC system, aiming to demonstrate IoFDT's feasibility and merits. Examples include DT-based collision prediction and avoidance system, and a DT-based car-sharing system, as typical and key Society 5.0 use-cases. Additionally, this platform can be utilized to develop a wide range of DT-based applications beyond smart mobility.

\section{Conclusion and Recommendations}
In this paper, we present the vision of IoFDT by covering its various aspects, including communication, learning, and computing, as well as implementation and platform. To advance IoFDT and facilitate Society 5.0, we provide several recommendations for advancing this field:

\begin{itemize}

\item \emph{Synchronization and Interaction}:
While individual DTs must be synchronized with physical system, in an IoFDT, it is also necessary to look at cross-DT interactions and coordination. 

\item \emph{Adaptive and Continuous Learning Frameworks}:
AI-native networks and learning frameworks that are capable of changing and adapting over time are needed to support the IoFDT. These frameworks must be responsive to new data and changes in physical systems to continuously enhance DTs' performance and reliability.

\item \emph{IoFDT-centric Resource Management}:
Due to the specific requirements of each DT in the IoFDT ecosystem, efficient allocation of network and computing resources, across the entire DT federation, is crucial for the IoFDT. This includes the implementation of learning-based dynamic network slicing and context-aware resource optimization.

\item \emph{Privacy and Security}:
Given the interconnected nature of IoFDT, it is crucial to ensure privacy-preserving insights across DTs and maintain system integrity and security by, e.g., leveraging federated intelligence, robust encryption protocols, and access control mechanisms in the IoFDT framework.

\item \emph{Application and Service}:
The development of novel applications and services enabled by IoFDT is necessary, and Society 5.0 offers fertile ground for this.

\end{itemize}

% \bibliographystyle{IEEEtran}
% \def\baselinestretch{0.82}%0.66
% \bibliography{./bib/walid,./bib/japan,./bib/omid}
% \vspace{-0.3cm}

% \section*{Biographies}
\begin{IEEEbiographynophoto}{Tao Yu}(yutao@mobile.ee.titech.ac.jp)
is a specially appointed Associate Professor with Tokyo Tech Academy for Super Smart Society at Tokyo Institute of Technology. His research interests include V2X, UAV communications, digital twin, and smart mobility. 
\end{IEEEbiographynophoto}

\begin{IEEEbiographynophoto}{Zongdian Li}(lizd@mobile.ee.titech.ac.jp) is an Assistant Professor in the Department of Electrical and Electronic Engineering, Tokyo Institute of Technology, Tokyo, Japan. His research interests include software-defined networking (SDN), vehicle-to-everything (V2X) communications, digital twins, and autonomous driving. 
\end{IEEEbiographynophoto}

\begin{IEEEbiographynophoto}{Omar Hashash}(omarnh@vt.edu)
is a Ph.D. student at the Electrical and Computer Engineering Department at Virginia Tech. His research interests include digital twins, the metaverse, wireless networks, and generalizable machine learning.
\end{IEEEbiographynophoto}

\begin{IEEEbiographynophoto}{Kei Sakaguchi}(sakaguchi@mobile.ee.titech.ac.jp)
is a Dean in Tokyo Tech Academy for Super Smart Society and a Professor at Tokyo Institute of Technology. His research interests include 5G/6G cellular networks, MIMO, wireless power transmission, autonomous driving, and super smart society.
\end{IEEEbiographynophoto}

\begin{IEEEbiographynophoto}{Walid Saad}(walids@vt.edu) is a Professor at the Department of Electrical and Computer Engineering at Virginia Tech and an IEEE Fellow. He works at the intersection of wireless networks, machine learning, and game theory. He received many major awards including the 2023 IEEE Marconi Prize Paper Award.
\end{IEEEbiographynophoto}

\begin{IEEEbiographynophoto}{M{\'e}rouane Debbah} (merouane.debbah@ku.ac.ae) is Professor at Khalifa University of Science and Technology, Abu Dhabi, UAE. He is an IEEE Fellow, a WWRF Fellow, a EurasipFellow, an AAIA Fellow, and an Institut Louis Bachelier Fellow. He has received more than 40 best paper awards.
 \end{IEEEbiographynophoto}

\end{document}